\documentclass[conference]{IEEEtran}
\IEEEoverridecommandlockouts
\usepackage{booktabs}
\usepackage{subcaption}
\usepackage{url} 
\usepackage{cite}
\usepackage{amsmath,amssymb,amsfonts}
\usepackage{algorithmic}
\usepackage{graphicx}
\usepackage{textcomp}
\usepackage{xcolor}
\def\BibTeX{{\rm B\kern-.05em{\sc i\kern-.025em b}\kern-.08em
    T\kern-.1667em\lower.7ex\hbox{E}\kern-.125emX}}
\begin{document}

\title{Dynamic Weight Adjusting Deep Q-Networks for Real-Time Environmental Adaptation
}

\author{\IEEEauthorblockN{1\textsuperscript{st} Xinhao Zhang}
\IEEEauthorblockA{\textit{Portland State University} \\
Portland, USA \\
xinhaoz@pdx.edu}
\and
\IEEEauthorblockN{2\textsuperscript{nd} Jinghan Zhang}
\IEEEauthorblockA{\textit{Portland State University} \\
Portland, USA \\
jinghanz@pdx.edu}
\and
\IEEEauthorblockN{3\textsuperscript{rd} Wujun Si}
\IEEEauthorblockA{\textit{Wichita State University} \\
Wichita, USA \\
wujun.si@wichita.edu}
\and
\IEEEauthorblockN{4\textsuperscript{th} Kunpeng Liu}
\IEEEauthorblockA{\textit{Portland State University} \\
Portland, USA \\
kunpeng@pdx.edu}

}

\maketitle

\begin{abstract}
Deep Reinforcement Learning  has shown excellent performance in generating efficient solutions for complex tasks. However, its efficacy is often limited by static training modes and heavy reliance on vast data from stable environments. To address these shortcomings, this study explores integrating dynamic weight adjustments into Deep Q-Networks (DQN) to enhance their adaptability. We implement these adjustments by modifying the sampling probabilities in the experience replay to make the model focus more on pivotal transitions as indicated by real-time environmental feedback and performance metrics. We design a novel \textit{Interactive Dynamic Evaluation Method} (IDEM) for DQN that successfully navigates dynamic environments by prioritizing significant transitions based on environmental feedback and learning progress. Additionally, when faced with rapid changes in environmental conditions, IDEM-DQN shows improved performance compared to baseline methods. Our results indicate that under circumstances requiring rapid adaptation, IDEM-DQN can more effectively generalize and stabilize learning. Extensive experiments across various settings confirm that IDEM-DQN outperforms standard DQN models, particularly in environments characterized by frequent and unpredictable changes.
\end{abstract}


\section{Introduction}

Deep Reinforcement Learning (DRL) has shown to be capable
of learning human-level control policies, outperforming traditional methods in domains from strategic game play to autonomous vehicle navigation~\cite{wang2022deep,urtans2018survey,li2017deep,arulkumaran2017deep}. Despite these advancements, the typical application of DRL, especially in the form of Deep Q-Networks (DQN), often falls short in dynamic environments where adaptability is crucial~\cite{saito2022simulation,cai2022survey,nie2023reinforcement,yang2024research,fan2020theoretical}. This limitation primarily stems from the static nature of traditional training methods and the substantial data requirements from relatively stable environments. Such conditions are not always feasible or realistic, particularly when environmental variables frequently change and unpredictability is high~\cite{yuan2022mec,han2022nrowan,9547816}.

The standard approach in DQN involves a fixed strategy for experience replay, which treats all transitions equally, regardless of their significance or the context of their occurrence~\cite{yang2020multi,wang2024research,wang2024machine,li2022improved}. While beneficial in stable settings, this method lacks the flexibility needed to cope with environments where the dynamics evolve more rapidly than the model can adapt~\cite{liu2022distributed}. Moreover, this approach may lead to inefficient learning and poor generalization when applied outside the training data distribution, as the model is often trained on a backlog of outdated or irrelevant experiences.

To address the issues found in DQN, researchers have developed several improvements, such as Double DQN~\cite{van2016deep,dan2024multiple,han2020research}, Prioritized Experience Replay~\cite{horgan2018distributed}, and Twin Delayed DDPG (TD3)~\cite{zhou2021novel,song2024looking,zhang2020td3}. Double DQN uses two networks to separate action selection from evaluation, reducing Q-value overestimations. However, this increases computational complexity and does not directly solve problems with low sample efficiency or insufficient exploration~\cite{huang2022intelligent}. Prioritized Experience Replay enhances learning efficiency by assigning different importance to each experience based on its Temporal Difference (TD) error. This method requires manual tuning of hyperparameters, which could introduce biases. TD3 reduces overestimations further with dual Q-networks and policy smoothing and enhances stability with delayed policy updates. However, TD3 is complex and challenging to tune, mainly suited for continuous action spaces~\cite{maurya2021federated}.

These approaches primarily focus on reducing overestimations and enhancing sample efficiency. Thus, these methods show limitations in dynamic or non-static environments. Although they improve training efficiency and bias reduction, they often fail to consider environmental changes, which leads to weak real-time adaptation. For instance, Double DQN and TD3 can stabilize the learning process and minimize Q-value overestimations with multiple networks and complex update mechanisms. However, their fixed network structures and update strategies might lack flexibility in rapidly changing environments. Additionally, while Prioritized Experience Replay optimizes data usage, its static priority settings and resampling mechanisms may not adjust quickly enough when the environment changes. This setting could harm the learning adaptability and outcomes.

Therefore, we think of introducing a dynamic adjustment to the experience replay mechanism in vanilla DQN. This replay mechanism in DQN influences the storing and reusing of past transitions (states, actions, rewards, new states) to enhance data utilization and stabilize learning. It breaks the correlation in time-series data through random sampling to help the network learn more effectively and stably. However, standard experience replay does not differentiate the importance of transitions, which might cause key experiences to be overlooked. Hence, we consider introducing a dynamic adjustment here for DQN to better adapt to environmental changes. We want this method to adjust the sampling frequency of important transitions based on current learning progress and environmental feedback. This improvement can enhance DQN's performance in various dynamic environments and make it more robust and effective in practical applications.

\textbf{Our Target.} To address these problems, we target three main aspects: 1) We aim to improve DQN's adaptability in dynamic environments by introducing a dynamic adjustment mechanism. This mechanism adjusts the sampling strategy or learning parameters in real time to enhance the model’s performance as conditions change. 2) We strive to make our improvements simple and stable. They should integrate seamlessly into the existing DQN architecture without adding too much complexity or computational load. This approach will enhance DQN’s generalization and robustness without losing efficiency. 3) We plan to test our method across various environments to evaluate DQN's performance and stability in real-world conditions.

\textbf{Our Method.} To achieve our goals, we design the \textbf{I}nteractive \textbf{D}ynamic \textbf{E}valuation \textbf{M}ethod (\textbf{IDEM}), an adaptive DQN adjustment framework optimized for dynamic environments. Specifically, to help DQN better adapt to environmental changes, we introduce dynamic weight adjustments into DQN's experience replay mechanism. We implement a mechanism that adjusts the importance of samples based on real-time environmental feedback. This involves modifying the weights of samples in the experience replay to prioritize those experiences that are most critical for improving the current strategy. Adjustments are based on the outcomes of actions; if an action yields better results than expected, we increase the replay probability for that type of action, and decrease it otherwise, focusing the learning process on transitions that could significantly enhance performance.

For our second goal, we design an adaptive learning rate adjustment function to ensure easy implementation and computational efficiency. We incorporate the learning rate adjustments based on performance feedback, effectively preventing overfitting and enhancing the model’s generalization capability. These designs can optimize DQN’s performance in dynamic environments while maintaining the algorithm's computational efficiency and stability.

Furthermore, our method addresses issues when the model encounters unknown or rare states. We achieve this through a dynamic adjustment mechanism that continually tweaks key learning parameters. This allows the model to adapt more quickly to new or infrequent environmental conditions, improving its responsiveness and adaptability. This dynamic adjustment strategy enables DQN to handle complex, dynamic environments without sacrificing learning efficiency.

In summary, our contribution includes:

\begin{enumerate}
    \item We propose a dynamic adjustment mechanism that uses real-time environmental feedback and performance-based importance adjustments of samples for DQN. This mechanism improves DQN's adaptability to dynamic environments and its sample efficiency.
    
    \item We create a simple yet effective weight adjustment function. This function includes adjustments to the learning rate and sample resampling probabilities. Thus, it allows us to optimize the learning strategy dynamically based on the model's immediate performance and feedback from the environment. This design in the mechanism boosts learning efficiency and enhances the model's generalization capabilities and robustness when facing new and rare states.
    
    \item We validate our method through a series of experiments. The results demonstrate significant advantages in enhancing DQN's ability to adapt to dynamic environments, optimizing learning efficiency, and increasing model stability. Compared to traditional DQN and its variants, our method shows better strategy learning and decision quality in various challenging environments.
\end{enumerate}

\section{Related Work and Preliminary}

DQN is a reinforcement learning method that integrates deep learning techniques with Q-learning, tailored for decision-making problems involving continuous state spaces and discrete action spaces~\cite{luong2019applications,alharin2020reinforcement,liu2024learning,zhu2023demonstration,padakandla2021survey,gao2016novel,liu2023meta}. In DQN, the state space $\mathcal{S}$ encompasses all possible states of the environment, and the action space $\mathcal{A}$ includes all possible actions an agent can 
execute~\cite{hafiz2022survey,steckelmacher2020sample,lemos2025enhancing,10.1145/3485447.3512083}. The policy $\pi$, typically parameterized by a neural network $Q(s, a; \theta)$, maps states $s \in \mathcal{S}$ to actions $a \in \mathcal{A}$, with $\theta$ representing the neural network parameters. The reward function $R(s, a)$ defines the immediate reward received by the agent after transitioning from state $s$ to a new state $s'$ through action $a$. The discount factor $\gamma$, generally set between 0 and 1, calculates the present value of future rewards, prioritizing nearer over more distant rewards.

The goal of DQN is to learn a policy $ \pi^* $ that maximizes the total expected return, i.e., the cumulative discounted rewards obtained by following policy $ \pi $ from any initial state $ s $. Mathematically, this is represented as:

\begin{equation}
    Q^*(s, a) = \mathbb{E} \left[ R(s, a) + \gamma \max_{a'} Q^*(s', a') \right],
\end{equation}

where $ Q^*(s, a) $ is the maximum expected return obtainable from taking action $ a $ in state $ s $.

In practical training, DQN iteratively updates the neural network parameters to approximate the optimal Q-function $ Q^*(s, a) $, using the update formula:

\begin{equation}
    \theta_{t+1} = \theta_t + \alpha \left[ y_t - Q(s_t, a_t; \theta_t) \right] \nabla_{\theta_t} Q(s_t, a_t; \theta_t),
\end{equation}

where $ y_t = r_t + \gamma \max_{a'} Q(s_{t+1}, a'; \theta_t) $ is the target Q-value, $ r_t $ is the immediate reward received, and $ \alpha $ is the learning rate. And our target is to find the optimal policy $ \pi^* $ that maximizes the expected cumulative reward, which can be mathematically expressed as:

\begin{equation}
    \pi^* = \arg \max_{\pi} \mathbb{E} \left[ \sum_{t=0}^{\infty} \gamma^t R(s_t, \pi(s_t)) \right],
\end{equation}

where $ \pi(s_t) $ denotes the action taken in state $ s_t $ according to policy $ \pi $, $ R(s_t, a_t) $ is the reward function, $ \gamma $ is the discount factor, representing the importance of future rewards.

\section{Methodology}
\subsection{Methodology Overview}

In the methodology section, we introduce the \textbf{I}nteractive \textbf{D}ynamic \textbf{E}valuation \textbf{M}ethod (\textbf{IDEM}), a strategy designed to address the adaptability issues and low sample efficiency of DQN in dynamic environments. We employ a dynamic weight adjustment mechanism to respond to environmental changes in real time and optimize the learning process. Additionally, we have developed an adaptive learning rate adjustment function to enhance the model's robustness and generalization capabilities when facing unfamiliar or rare states. Our methodology is divided into three main parts: (1) \textbf{Dynamic Weight Adjustment Mechanism}: We describe how we adjust the weights of samples in the experience replay based on environmental feedback and the progress of learning with IDEM. In this part, we aim to guide the model to focus more on significant transitions that are crucial for improving current strategies; (2) \textbf{Adaptive Learning Rate Adjustment Function}: We introduce and analyze the traction function of the IDEM mechanism, as well as explore its mathematical properties and impact within DQN. This traction function dynamically adjusts the learning rate based on the immediate performance of the model. In this part, the function helps the model to better adapt to the complexities and variabilities of the training environment; (3) \textbf{Implementation of the IDEM}: We detail how we integrate the IDEM method with DQN, and explain the practical steps and modifications made to the standard DQN framework to incorporate our dynamic adjustments.

\begin{figure}
    
    \centering
    \includegraphics[width=0.9\linewidth]{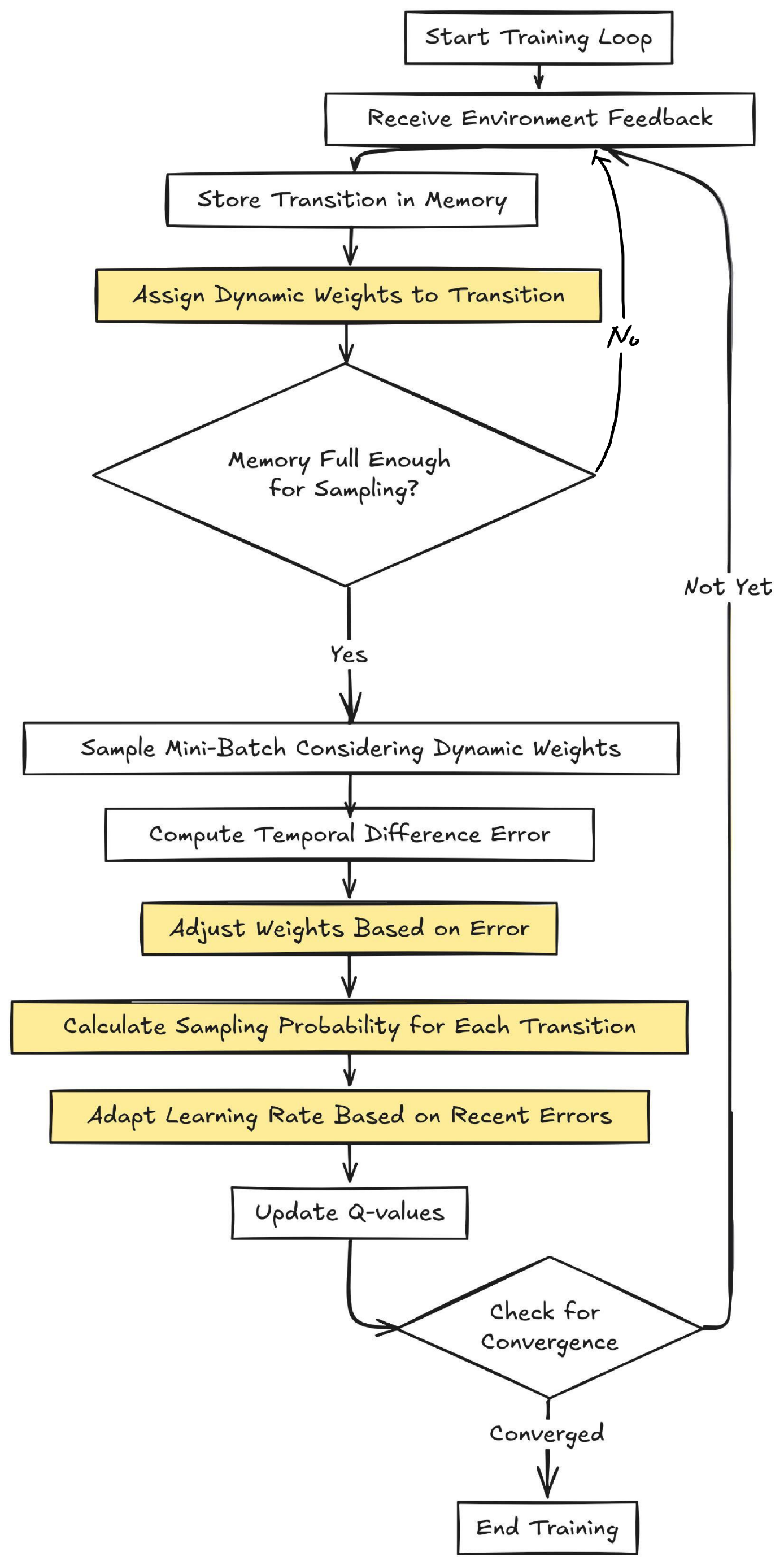}
    \caption{Flowchart of IDEM-DQN. Here the IDEM focuses on assigning weights to transitions based on their significance and adjusting learning rates based on real-time error metrics. These enhancements prioritize crucial learning opportunities and optimize the model's response to changing environmental conditions.}
    \label{fig:mermaid}
    
\end{figure}

\subsection{Dynamic Weight Adjustment Mechanism}
Here, we aim to enhance the model's adaptability and learning efficiency in dynamic environments. Therefore, we design a mechanism for dynamically adjusting the weights of experiences within the replay buffer. This mechanism prioritizes learning from the most crucial transitions at any given point in the learning process by adjusting the weights of experiences based on their relevance and impact.

The experience replay mechanism in DQN traditionally stores transitions $ (s_t, a_t, r_t, s_{t+1}) $ from agent-environment interactions. These transitions are crucial for updating Q-value estimates to optimize the policy. Our goal is to significantly enhance DQN's learning efficiency and adaptability, particularly in dynamic environments where the relevance of experiences can change over time. To achieve this, the IDEM introduces a dynamic weighting system into the experience replay. This system aims to amplify the impact of significant transitions on the learning process. In the IDEM framework, the dynamic weighting mechanism is applied to the experience replay component of DQN. Each transition in the replay buffer, denoted by $ (s_t, a_t, r_t, s_{t+1}) $, is associated with a dynamic weight $ w_t $ that influences how frequently it is sampled for training the network. In this way, it ensures that more informative experiences exert a greater influence on model updates.

\paragraph{Temporal Difference Error Calculation} Specifically, the TD error for a transition is fundamental in updating the weights and is calculated using the following equation:

\begin{equation}
    \delta_t = r_t + \gamma \max_{a'} Q(s_{t+1}, a'; \theta) - Q(s_t, a_t; \theta),
\end{equation}

where $r_t$ is the reward received after taking action $a_t$ in state $s_t$, $\gamma$ is the discount factor representing the importance of future rewards, $Q(s_{t+1}, a'; \theta)$ estimates the maximum future reward from the next state $s_{t+1}$, and $\theta$ are the parameters of the Q-network. This equation reflects the difference between the predicted reward for the action taken in the current state and the maximum predicted reward for the best possible action in the next state, adjusted by the immediate reward received and discounted by $\gamma$. The magnitude of $\delta_t$ indicates the degree to which the current Q-value predictions are off-target, which in turn informs how much the model needs to adjust its predictions. A high absolute value of $\delta_t$ suggests a significant error in prediction, implying that the learning from this particular transition could lead to substantial improvements in policy performance.

\paragraph{Weight Adjustment Formula} The weight $ w_t $ of each transition in the experience replay is dynamically adjusted based on the absolute temporal difference error:
\begin{equation}
    w_t = \exp(\lambda |\delta_t|),
\end{equation}

where $\lambda$ is a positive scaling factor that determines how sensitive the weight adjustments are to the TD error. This formula uses the exponential function to scale the weights of transitions in the experience replay, with the scaling factor $\lambda$ modulating the extent of the adjustment. A larger $\lambda$ increases the responsiveness of the weight to changes in the TD error, enhancing the priority of transitions with larger errors during training. This method prioritizes transitions that the model currently mispredicts the most, under the assumption that these transitions are more informative and, thus, more valuable for learning. The exponential function ensures that even small discrepancies are noted but gives exponentially greater importance to larger discrepancies, thereby focusing learning efforts where they are needed most.

\paragraph{Sampling Mechanism Based on Updated Weights} The probability of sampling a particular transition from the experience replay buffer is adjusted in proportion to its weight:
\begin{equation}
    P(t) = \frac{w_t}{\sum_{i} w_i},
\end{equation}
where $ P(t) $ is the probability of sampling transition $ t $, $ w_t $ is the weight of transition $ t $, and $ \sum_{i} w_i $ is the sum of weights for all transitions stored in the replay buffer. This formula increases the frequency of sampling transitions with higher weights during training. We calculate the sampling probability $ P(t) $ by weighting it with $ w_t $, the transition's weight. This method allows us to focus on transitions with substantial prediction errors, thus prioritizing the most crucial learning opportunities. We normalize the weights by the sum of all weights, $ \sum_{i} w_i $, to ensure that $ P(t) $ forms a valid probability distribution. This normalization keeps the sampling process fair and balanced while focusing on the most valuable experiences.

\subsection{Adaptive Learning Rate Adjustment Function}
In reinforcement learning with DQN, it is essential to manage the source of learning experiences and the pace at which learning unfolds. Traditional DQN often employs a constant learning rate, which might not cope well with the variable complexities and demands of different environments. A static rate could be too slow for simple scenarios or too fast in complex or volatile conditions.

\paragraph{Adaptive Learning Rate Adjustment Function} To overcome these issues, we integrate an \textbf{\textit{Adaptive Learning Rate Adjustment Function}} in Eq.~\ref{eq:adjustfunc} in the IDEM framework. This function adjusts the learning rate $ \eta_t $ of the Q-network to match the precision of the model's predictions. By fine-tuning $ \eta_t $ according to recent performance metrics, it customizes the learning speed to the current environmental conditions, thus enhancing both learning efficiency and stability:
\begin{equation}
    \eta_t = \eta_0 \cdot \exp(-\kappa \cdot \overline{\delta}_t),
    \label{eq:adjustfunc}
\end{equation}

where $ \eta_t $ is the learning rate at time $ t $, $ \eta_0 $ is the initial learning rate, $ \kappa $ is a positive decay factor, and $ \overline{\delta}_t $ represents the moving average of the absolute temporal difference errors over a recent window of transitions. This function adapts the learning rate based on the magnitude of recent errors; larger average errors suggest significant prediction discrepancies and necessitate a slower learning rate for stable convergence, while smaller errors indicate more accurate predictions and allow for an increased learning rate to expedite convergence.
\paragraph{Continuity and Differentiability in Backpropagation}
Continuity. The adaptive learning rate function is continuous for all $x$, and $-\kappa \cdot \overline{\delta}_t$ is a linear transformation of the moving average of absolute temporal difference errors, also assumed continuous if $\overline{\delta}_t$ is derived from real-valued data sequences. Hence, the overall function $\eta_t$ is continuous. This continuity ensures that small changes in the temporal difference errors lead to minor adjustments in the learning rate, thus providing stability and smooth responses to changes in training dynamics.

Differentiability. Furthermore, the function $\eta_t$ is differentiable where its components are differentiable. Since the exponential function $\exp(x)$ is differentiable for all real numbers $x$, and the term $-\kappa \cdot \overline{\delta}_t$ is a linear (and hence differentiable) transformation, it follows that $\eta_t$ is differentiable as well. This property is crucial for enabling gradient-based optimization techniques such as backpropagation to compute gradients and update model parameters effectively, avoiding issues like discontinuities or non-differentiable points.

According to the analysis, we see that:
\begin{itemize}
    \item The differentiable and continuous nature of $\eta_t$ aligns well with the gradient descent steps in backpropagation, where precise control over the learning rate is essential for achieving convergence. By smoothly adjusting based on the error metrics, $\eta_t$ allows the gradient descent algorithm to leverage this dynamic adjustment without interruptions.
    \item The function's dependence on $\overline{\delta}_t$ enables a responsive adaptation of the learning rate to the actual performance and error landscape of the model. This adjustability prevents scenarios where the learning rate might be too high, causing overshooting, or too low, leading to excessively slow convergence. This dynamic adjustment is essential for maintaining robust performance in environments where the characteristics and challenges can change unpredictably.
\end{itemize}

\section{Experiments}

In the experimental section of our study, we conduct a series of tests in a standard testing environment \texttt{FrozenLake} 
to evaluate the performance of IDEM-DQN compared to standard DQN. We aim to answer three key questions through the following experiments:
\begin{enumerate}
    \item How does the performance of IDEM-DQN compare to that of standard DQN in a static environment?
    \item How does IDEM-DQN perform relative to vanilla DQN in dynamically changing environments?
    \item What is the impact of IDEM's parameters on the effectiveness of IDEM-DQN?
\end{enumerate}

To answer these questions, we design the following experiments: (1) Baseline Performance Comparison. We first establish a baseline by running both the IDEM-DQN and standard DQN on the classic \texttt{FrozenLake}\footnote{\url{https://www.gymlibrary.dev/environments/toy_text/frozen_lake/}}. We measure and compare the average cumulative rewards, the number of episodes to convergence, and the stability of the learning curves; (2) Adaptability in Changing Environments. We test the adaptability of IDEM-DQN by introducing sudden changes in the environment dynamics; (3) Sensitivity to Parameter Settings. We conduct a sensitivity analysis on the parameters learning rate and Adam's $\beta_1$, which control the step size of updates and the exponential decay rate for the first moment estimates of past gradients, respectively. $\beta_1$ can be seen as the model's short-term memory of gradient changes. We vary these parameters and observe their impact on average winning steps, win rate, average reward, and average loss. We use a grid search approach to systematically explore a range of values for each parameter.


\subsection{Experimental Settings}

\paragraph{Environment} 
In this study, we select the standard \texttt{FrozenLake} environment from AI Gym, which is a simple, grid-based maze challenge. The grid is composed of a start state $S$ and a goal state $G$, where the main objective is navigating from $S$ to $G$ across a simulated frozen lake. The grid includes ``frozen'' states marked $F$ which are safe for traversal, and ``hole'' states marked $H$ that terminate the episode upon entry. Agents can perform one of four actions: left ($L$), down ($D$), right ($R$), or up ($U$). Actions leading off the grid result in the agent remaining in its current position.

This environment's dynamics introduce an element of unpredictability through the slipperiness of the ice, affecting movement decisions. For example, selecting action $L$ might actually result in movement in any of the directions $\{U, L, D\}$ with an equal probability of $\frac{1}{3}$. This stochasticity extends similarly across other actions. The reward function $R(s, a)$ grants a reward of 1 if the state $s$ is the goal $G$, and 0 otherwise. The game ends once the agent enters a termination state from the set $\{H, G\}$, where it then remains indefinitely. This environment setup tests the agent’s navigation skills under conditions that mirror the unpredictability found in real-world scenarios, requiring adaptive and strategic responses.
\paragraph{Inplemental Details} In this study, we configured our experiments using a custom setup in the \texttt{FrozenLake-v1} with a 4x4 or 8x8 grid. The DQN part comprises two layers: the first layer maps the state space to $50$ hidden units, and the second maps these hidden units to the number of possible actions. The network parameters are optimized using the Adam optimizer with a learning rate of $0.0001$.

The agent's behavior is controlled by an epsilon-greedy strategy, with $\epsilon$ set to $0.1$. This setup is to balance between exploration and exploitation during the training process. We employ a memory buffer capable of storing up to $3000$ transitions to maintain an essential balance between recent and past experiences. This buffer is crucial for our dynamic weighting mechanism.

For our experimental runs, we set the batch size for network training to $1000$ and conducted a total of $3000$ episodes to sufficiently train and evaluate our model. These settings ensure rigorous testing and validation of the modified DQN model under controlled, reproducible conditions.
\subsection{Overall Performance}
In this experiment, we aim to evaluate and compare the performance of the standard DQN model and our IDEM-DQN model. Here we measure the average number of steps required to successfully reach the goal and observe the changes in loss across epochs in a 4x4 grid.

\begin{figure}[h]
    \centering
    \begin{subfigure}[b]{0.49\linewidth} 
        \centering
        \includegraphics[width=\linewidth]{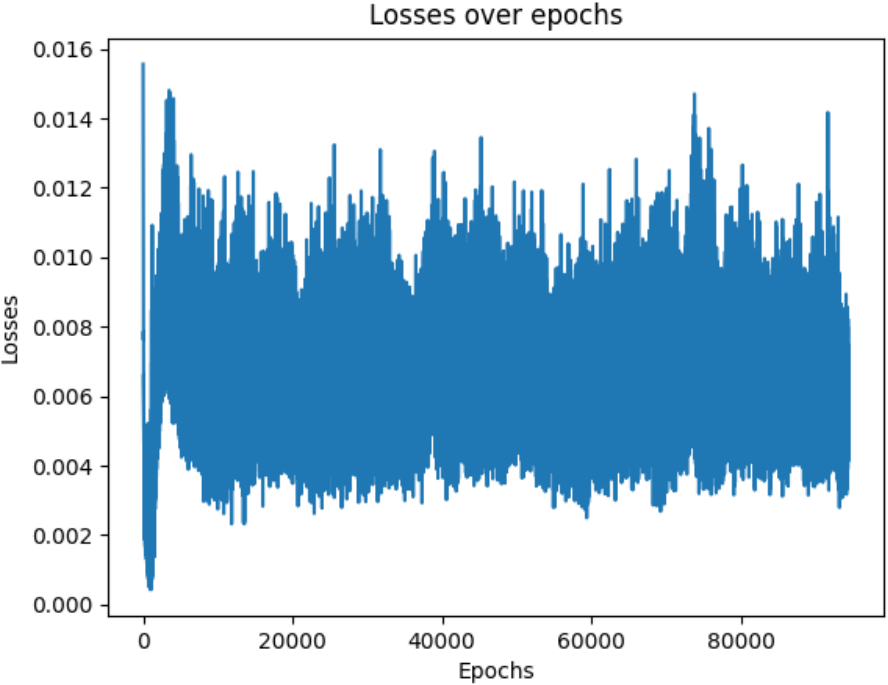} 
        \caption{DQN loss.}
        \label{fig:dqn}
    \end{subfigure}
    \hfill 
    \begin{subfigure}[b]{0.49\linewidth} 
        \centering
        \includegraphics[width=\linewidth]{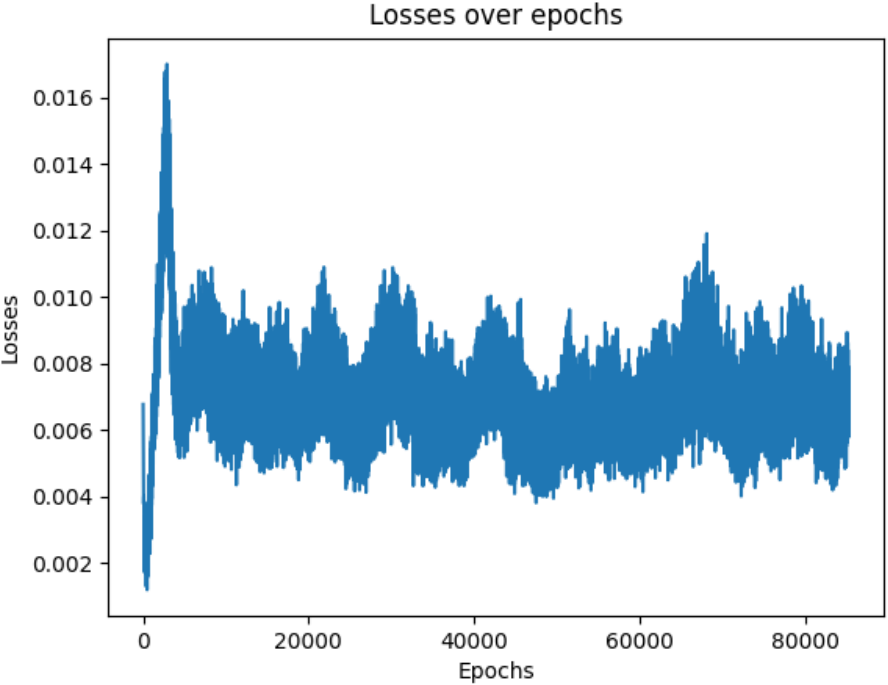} 
        \caption{IDEM-DQN loss.}
        \label{fig:idem_dqn}
    \end{subfigure}
    \caption{Loss comparison of DQN and IDEM-DQN on 4x4 FrozenLake grid.}
    \label{fig:loss_comparison}
\end{figure}

As shown in Fig.~\ref{fig:dqn} and Fig.~\ref{fig:idem_dqn}, the trends of losses for both standard and IDEM-DQN models fluctuate but generally decrease. Initially, both models experience peaks in loss as they adapt to the environment. Then, the IDEM-DQN model shows a smoother and quicker stabilization in loss reduction. This demonstrates that IDEM-DQN handles environmental complexities better due to its dynamic adjustment capabilities.

Moreover, the results in TABLE~\ref{tab:winsteps} and TABLE~\ref{tab:88} demonstrate that IDEM-DQN consistently outperforms the traditional DQN across all metrics. IDEM-DQN shows lower average winning steps, higher win rates, and higher rewards in both 4x4 and 8x8 environments. These improvements highlight the adaptability and effectiveness of the IDEM-DQN under more dynamically changing conditions.

  

\begin{table}[htbp]
  \centering
  \caption{Average Winning Steps comparison of DQN and IDEM-DQN on the 4x4 FrozenLake (Lower values indicate better performance). The best results are highlighted in \textbf{bold}.}
  \label{tab:winsteps}
  \begin{tabular}{lcc}
    \toprule
    Metric                     & DQN   & IDEM-DQN \\
    \midrule
    Average Winning Steps      & 35    & \textbf{33.35} \\
    \bottomrule
  \end{tabular}

\end{table}

\begin{figure}[h]
    \centering
    \includegraphics[width=0.45\textwidth]{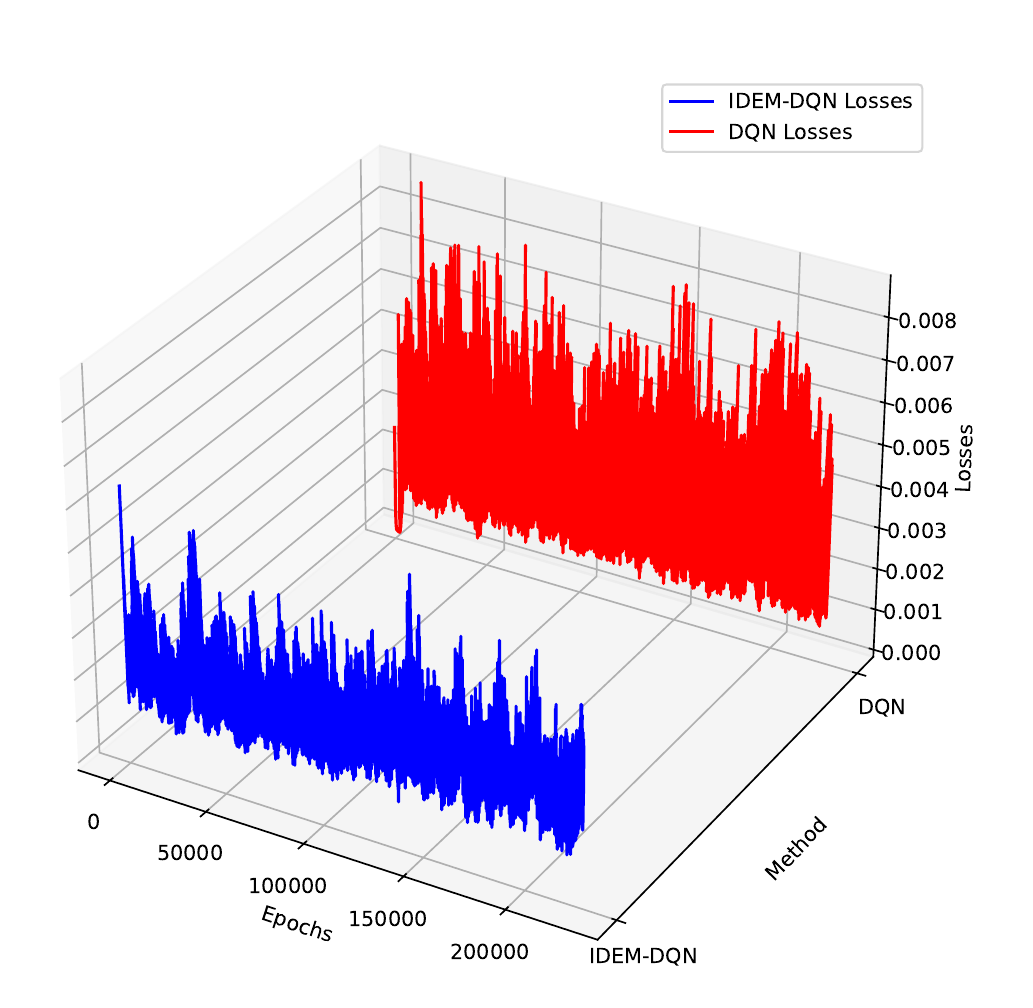}
    \caption{DQN and IDEM-DQN performance in 8x8 FrozenLake grid.}
    \label{fig:losses}
\end{figure}

\begin{table}[h]
    \centering
    \caption{DQN and IDEM-DQN performance in an 8x8 FrozenLake environment. The best results are highlighted in \textbf{bold}.}
    \label{tab:88}
    \begin{tabular}{lcc}
    \toprule
    \textbf{Metric}               & \textbf{DQN} & \textbf{IDEM-DQN} \\
    \midrule
    Average Winning Steps         & 91.34        & \textbf{88.54}    \\
    Win rate                      & 0.41         & \textbf{0.42}     \\
    Average reward                & 0.41         & \textbf{0.42}     \\
    \bottomrule
    \end{tabular}

\end{table}
\begin{figure*}[htbp]
    \centering  
    \includegraphics[width=0.85\textwidth]{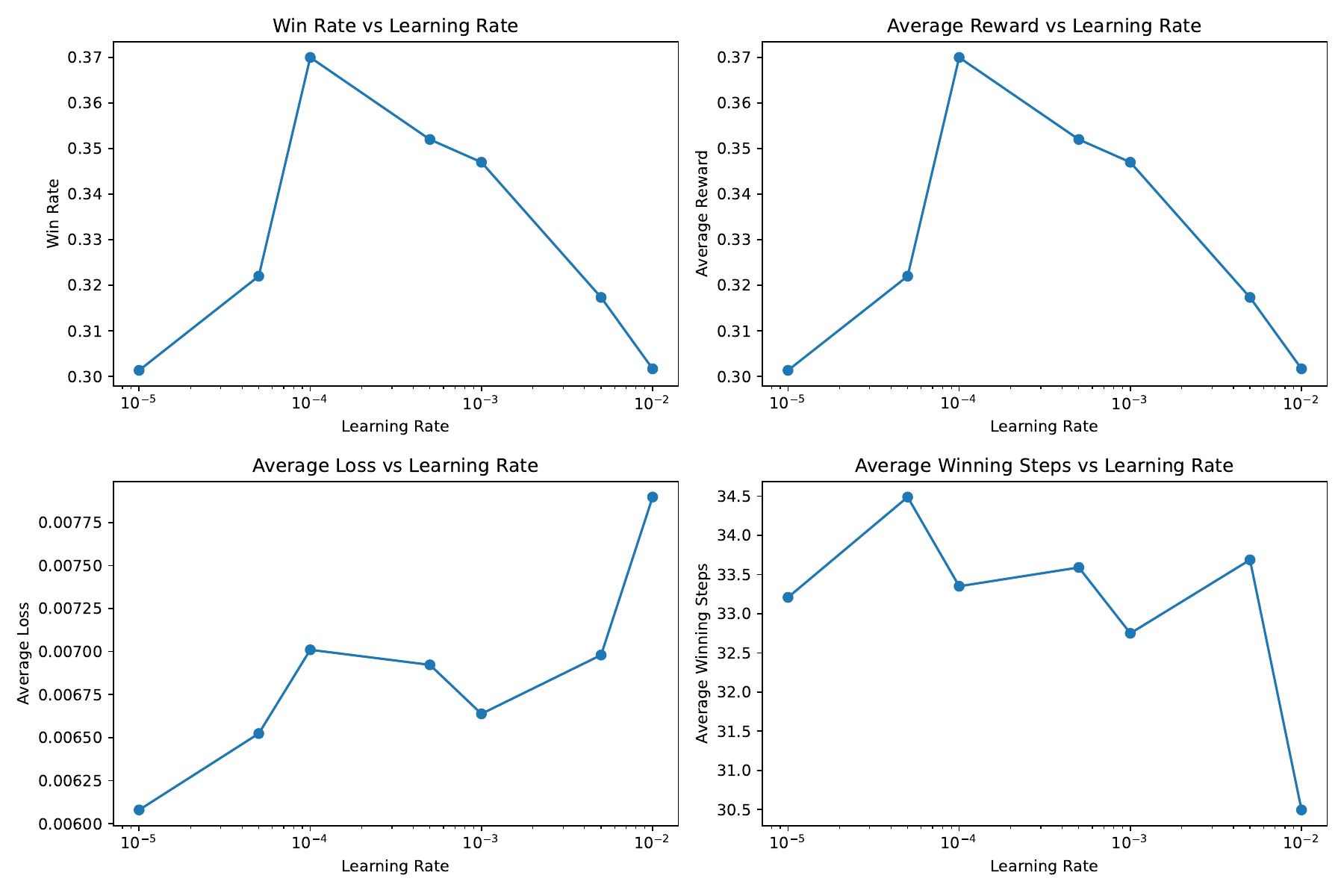}
    \caption{Ablation study of the learning rate.}
    \label{fig:ablation_study_lr}

    \includegraphics[width=0.85\textwidth]{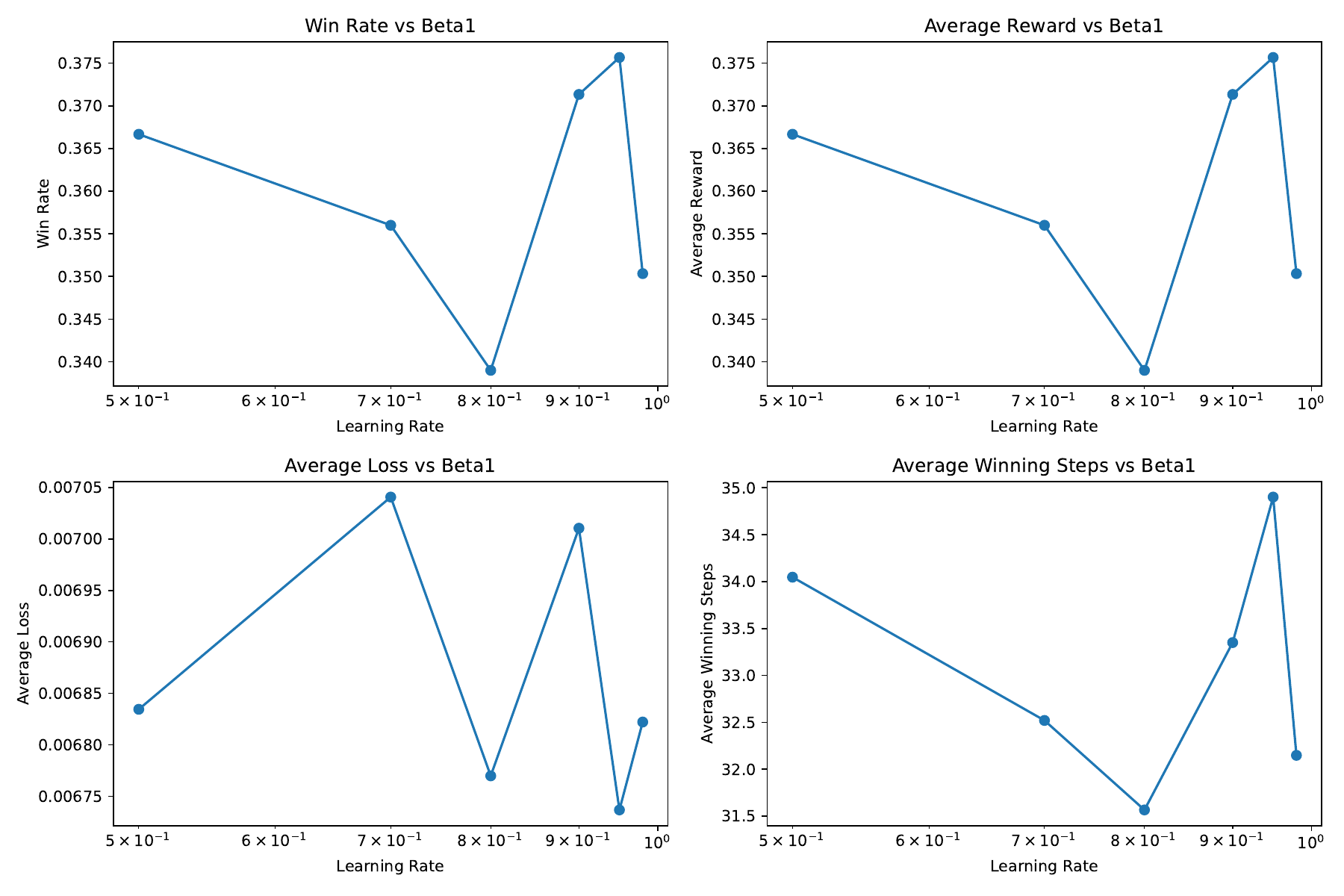}
    \caption{Ablation study of $\beta_1$}
    \label{fig:ablation_study_beta1}
    
\end{figure*}

\subsection{Adaptive Performance in Dynamic Environments}
In this experiment, we aim to assess the adaptive capabilities of both the DQN and IDEM-DQN models in response to dynamically changing conditions in the grid. In this experiment, we randomly alter the stability of the ice tiles and periodically move the goal location to create a more challenging and variable scenario for the models. In this dynamic setting, we evaluate the speed at which each model adapts to these changes, focusing on metrics such as the time required to return to high performance and the variability in episode rewards. The changes are introduced at random intervals to thoroughly test each model's ability to re-learn and adjust strategies quickly.
  

\begin{table}[htbp]
  \centering
  \caption{Performance comparison of DQN and IDEM-DQN in dynamic environments. The best results are highlighted in \textbf{bold}.}
  \label{tab:dynenv}
    \begin{tabular}{lcc}
    \toprule
    \textbf{Metric}                    & \textbf{DQN}             & \textbf{IDEM-DQN} \\
    \midrule
    Average Winning Steps     & 9.93                    & \textbf{7.88} \\
    Win rate                  & 0.83                    & \textbf{0.88} \\
    Average reward            & 0.83                    & \textbf{0.88} \\
    Average Loss              & $1.73 \times 10^{-4}$   & $\mathbf{1.39442 \times 10^{-4}}$ \\
    \bottomrule
    \end{tabular}%
  
\end{table}

The results, as summarized in Fig.~\ref{fig:losses} 
and TABLE~\ref{tab:dynenv}, reveal that IDEM-DQN significantly outperforms the standard DQN model across all metrics. The IDEM-DQN model demonstrates quicker adaptation to changes with lower average winning steps, higher win rates, and improved average rewards. Additionally, it maintains a lower average loss, indicating more stable and efficient learning under dynamic conditions. These findings confirm the enhanced robustness and adaptability of the IDEM-DQN model to work in environments where conditions frequently change.

\subsection{Ablation Study}
In ablation study, we conduct an ablation study to investigate the influence of learning rate and the $\beta_1$ parameter on the effectiveness of the IDEM-DQN model. As shown in Fig.~\ref{fig:ablation_study_lr} and Fig.~\ref{fig:ablation_study_beta1}, we highlight significant variations in performance metrics with changes in these parameters. For the learning rate, the optimal range is found between $10^{-4}$ and $10^{-3}$, where the model achieves higher win rates and average rewards coupled with a decrease in average loss. This suggests that a mid-range learning rate effectively balances the model's ability to adapt quickly without overwhelming the learning process with rapid updates. Regarding the $\beta_1$ parameter, the figures show that values around $0.8$ to $0.9$ optimize win rates and rewards. This suggests that a higher momentum term in the Adam optimizer improves IDEM-DQN's navigation through its learning landscape. However, the variable average loss across different $\beta_1$ values indicates the model's sensitivity to this parameter. Thus $\beta_1$ is vital to ensure stability and prevent performance degradation due to overfitting or erratic learning behavior.


\section{Conclusion}

In conclusion, our study introduces a novel method, IDEM, to enhance DQN's adaptability in dynamic environments through dynamic weight adjustments and adaptive learning rate modifications. IDEM-DQN prioritizes significant transitions based on real-time feedback and outperforms standard DQN models, particularly in environments with frequent and unpredictable changes. We conducted a series of extensive tests to confirm that these strategic modifications improve learning efficiency and model robustness without the computational complexity of more advanced models. The results demonstrate that IDEM-DQN is an effective and scalable solution for real-world applications where environmental conditions rapidly evolve, and that targeted enhancements to DQN frameworks can significantly boost performance in dynamic settings.

\section*{Acknowledgment}

The authors Jinghan Zhang, Xinhao Zhang and Kunpeng Liu are supported by NSF 2348485 and NSF 2426339.

\vspace{12pt}

\bibliographystyle{ieeetr}
\bibliography{ref}

\begin{thebibliography}{10}

\bibitem{wang2022deep}
X.~Wang, S.~Wang, X.~Liang, D.~Zhao, J.~Huang, X.~Xu, B.~Dai, and Q.~Miao, ``Deep reinforcement learning: A survey,'' {\em IEEE Transactions on Neural Networks and Learning Systems}, vol.~35, no.~4, pp.~5064--5078, 2022.

\bibitem{urtans2018survey}
E.~Urtans and A.~Nikitenko, ``Survey of deep q-network variants in pygame learning environment,'' in {\em Proceedings of the 2018 2nd International Conference on Deep Learning Technologies}, pp.~27--36, 2018.

\bibitem{li2017deep}
Y.~Li, ``Deep reinforcement learning: An overview,'' {\em arXiv preprint arXiv:1701.07274}, 2017.

\bibitem{arulkumaran2017deep}
K.~Arulkumaran, M.~P. Deisenroth, M.~Brundage, and A.~A. Bharath, ``Deep reinforcement learning: A brief survey,'' {\em IEEE Signal Processing Magazine}, vol.~34, no.~6, pp.~26--38, 2017.

\bibitem{saito2022simulation}
N.~Saito, T.~Oda, A.~Hirata, K.~Toyoshima, M.~Hirota, and L.~Barolli, ``Simulation results of a dqn based aav testbed in corner environment: a comparison study for normal dqn and tls-dqn,'' in {\em Innovative Mobile and Internet Services in Ubiquitous Computing: Proceedings of the 15th International Conference on Innovative Mobile and Internet Services in Ubiquitous Computing (IMIS-2021)}, pp.~156--167, Springer, 2022.

\bibitem{cai2022survey}
Q.~Cai, C.~Cui, Y.~Xiong, W.~Wang, Z.~Xie, and M.~Zhang, ``A survey on deep reinforcement learning for data processing and analytics,'' {\em IEEE Transactions on Knowledge and Data Engineering}, vol.~35, no.~5, pp.~4446--4465, 2022.

\bibitem{nie2023reinforcement}
M.~Nie, D.~Chen, and D.~Wang, ``Reinforcement learning on graphs: A survey,'' {\em IEEE Transactions on Emerging Topics in Computational Intelligence}, vol.~7, no.~4, pp.~1065--1082, 2023.

\bibitem{yang2024research}
Q.~Yang, Z.~Wang, S.~Liu, and Z.~Li, ``Research on improved u-net based remote sensing image segmentation algorithm,'' {\em arXiv preprint arXiv:2408.12672}, 2024.

\bibitem{fan2020theoretical}
J.~Fan, Z.~Wang, Y.~Xie, and Z.~Yang, ``A theoretical analysis of deep q-learning,'' in {\em Learning for dynamics and control}, pp.~486--489, PMLR, 2020.

\bibitem{yuan2022mec}
X.~Yuan, H.~Tian, Z.~Zhang, Z.~Zhao, L.~Liu, A.~K. Sangaiah, and K.~Yu, ``A mec offloading strategy based on improved dqn and simulated annealing for internet of behavior,'' {\em ACM Transactions on Sensor Networks}, vol.~19, no.~2, pp.~1--20, 2022.

\bibitem{han2022nrowan}
S.~Han, W.~Zhou, J.~Lu, J.~Liu, and S.~L{\"u}, ``Nrowan-dqn: A stable noisy network with noise reduction and online weight adjustment for exploration,'' {\em Expert Systems with Applications}, vol.~203, p.~117343, 2022.

\bibitem{9547816}
K.~Liu, Y.~Fu, L.~Wu, X.~Li, C.~Aggarwal, and H.~Xiong, ``Automated feature selection: A reinforcement learning perspective,'' {\em IEEE Transactions on Knowledge and Data Engineering}, vol.~35, no.~3, pp.~2272--2284, 2023.

\bibitem{yang2020multi}
Y.~Yang, L.~Juntao, and P.~Lingling, ``Multi-robot path planning based on a deep reinforcement learning dqn algorithm,'' {\em CAAI Transactions on Intelligence Technology}, vol.~5, no.~3, pp.~177--183, 2020.

\bibitem{wang2024research}
Z.~Wang, H.~Yan, C.~Wei, J.~Wang, S.~Bo, and M.~Xiao, ``Research on autonomous driving decision-making strategies based deep reinforcement learning,'' {\em arXiv preprint arXiv:2408.03084}, 2024.

\bibitem{wang2024machine}
M.~Wang and S.~Liu, ``Machine learning-based research on the adaptability of adolescents to online education,'' {\em arXiv preprint arXiv:2408.16849}, 2024.

\bibitem{li2022improved}
J.~Li, Y.~Chen, X.~Zhao, and J.~Huang, ``An improved dqn path planning algorithm,'' {\em The Journal of Supercomputing}, vol.~78, no.~1, pp.~616--639, 2022.

\bibitem{liu2022distributed}
S.~Liu and M.~Zhu, ``Distributed inverse constrained reinforcement learning for multi-agent systems,'' {\em Advances in Neural Information Processing Systems}, vol.~35, pp.~33444--33456, 2022.

\bibitem{van2016deep}
H.~Van~Hasselt, A.~Guez, and D.~Silver, ``Deep reinforcement learning with double q-learning,'' in {\em Proceedings of the AAAI conference on artificial intelligence}, vol.~30, 2016.

\bibitem{dan2024multiple}
H.-C. Dan, P.~Yan, J.~Tan, Y.~Zhou, and B.~Lu, ``Multiple distresses detection for asphalt pavement using improved you only look once algorithm based on convolutional neural network,'' {\em International Journal of Pavement Engineering}, vol.~25, no.~1, p.~2308169, 2024.

\bibitem{han2020research}
B.-A. Han and J.-J. Yang, ``Research on adaptive job shop scheduling problems based on dueling double dqn,'' {\em Ieee Access}, vol.~8, pp.~186474--186495, 2020.

\bibitem{horgan2018distributed}
D.~Horgan, J.~Quan, D.~Budden, G.~Barth-Maron, M.~Hessel, H.~Van~Hasselt, and D.~Silver, ``Distributed prioritized experience replay,'' {\em arXiv preprint arXiv:1803.00933}, 2018.

\bibitem{zhou2021novel}
J.~Zhou, S.~Xue, Y.~Xue, Y.~Liao, J.~Liu, and W.~Zhao, ``A novel energy management strategy of hybrid electric vehicle via an improved td3 deep reinforcement learning,'' {\em Energy}, vol.~224, p.~120118, 2021.

\bibitem{song2024looking}
Y.~Song, P.~Arora, S.~T. Varadharajan, R.~Singh, M.~Haynes, and T.~Starner, ``Looking from a different angle: Placing head-worn displays near the nose,'' in {\em Proceedings of the Augmented Humans International Conference 2024}, pp.~28--45, 2024.

\bibitem{zhang2020td3}
F.~Zhang, J.~Li, and Z.~Li, ``A td3-based multi-agent deep reinforcement learning method in mixed cooperation-competition environment,'' {\em Neurocomputing}, vol.~411, pp.~206--215, 2020.

\bibitem{huang2022intelligent}
L.~Huang, M.~Ye, X.~Xue, Y.~Wang, H.~Qiu, and X.~Deng, ``Intelligent routing method based on dueling dqn reinforcement learning and network traffic state prediction in sdn,'' {\em Wireless Networks}, pp.~1--19, 2022.

\bibitem{maurya2021federated}
S.~Maurya, S.~Joseph, A.~Asokan, A.~A. Algethami, M.~Hamdi, H.~T. Rauf, {\em et~al.}, ``Federated transfer learning for authentication and privacy preservation using novel supportive twin delayed ddpg (s-td3) algorithm for iiot,'' {\em Sensors}, vol.~21, no.~23, p.~7793, 2021.

\bibitem{luong2019applications}
N.~C. Luong, D.~T. Hoang, S.~Gong, D.~Niyato, P.~Wang, Y.-C. Liang, and D.~I. Kim, ``Applications of deep reinforcement learning in communications and networking: A survey,'' {\em IEEE communications surveys \& tutorials}, vol.~21, no.~4, pp.~3133--3174, 2019.

\bibitem{alharin2020reinforcement}
A.~Alharin, T.-N. Doan, and M.~Sartipi, ``Reinforcement learning interpretation methods: A survey,'' {\em IEEE Access}, vol.~8, pp.~171058--171077, 2020.

\bibitem{liu2024learning}
S.~Liu and M.~Zhu, ``Learning multi-agent behaviors from distributed and streaming demonstrations,'' {\em Advances in Neural Information Processing Systems}, vol.~36, 2024.

\bibitem{zhu2023demonstration}
Y.~Zhu, C.~Honnet, Y.~Kang, J.~Zhu, A.~J. Zheng, K.~Heinz, G.~Tang, L.~Musk, M.~Wessely, and S.~Mueller, ``Demonstration of chromocloth: Re-programmable multi-color textures through flexible and portable light source,'' in {\em Adjunct Proceedings of the 36th Annual ACM Symposium on User Interface Software and Technology}, pp.~1--3, 2023.

\bibitem{padakandla2021survey}
S.~Padakandla, ``A survey of reinforcement learning algorithms for dynamically varying environments,'' {\em ACM Computing Surveys (CSUR)}, vol.~54, no.~6, pp.~1--25, 2021.

\bibitem{gao2016novel}
H.~Gao, H.~Wang, Z.~Feng, M.~Fu, C.~Ma, H.~Pan, B.~Xu, and N.~Li, ``A novel texture extraction method for the sedimentary structures’ classification of petroleum imaging logging,'' in {\em Pattern Recognition: 7th Chinese Conference, CCPR 2016, Chengdu, China, November 5-7, 2016, Proceedings, Part II 7}, pp.~161--172, Springer, 2016.

\bibitem{liu2023meta}
S.~Liu and M.~Zhu, ``Meta inverse constrained reinforcement learning: Convergence guarantee and generalization analysis,'' in {\em The Twelfth International Conference on Learning Representations}, 2023.

\bibitem{hafiz2022survey}
A.~Hafiz, ``A survey of deep q-networks used for reinforcement learning: state of the art,'' {\em Intelligent Communication Technologies and Virtual Mobile Networks: Proceedings of ICICV 2022}, pp.~393--402, 2022.

\bibitem{steckelmacher2020sample}
D.~Steckelmacher, H.~Plisnier, D.~M. Roijers, and A.~Now{\'e}, ``Sample-efficient model-free reinforcement learning with off-policy critics,'' in {\em Machine Learning and Knowledge Discovery in Databases: European Conference, ECML PKDD 2019, W{\"u}rzburg, Germany, September 16--20, 2019, Proceedings, Part III}, pp.~19--34, Springer, 2020.

\bibitem{lemos2025enhancing}
M.~L. H.~D. Lemos, A.~R. Tavares, L.~S. Marcolino, L.~Chaimowicz, {\em et~al.}, ``Enhancing deep reinforcement learning for scale flexibility in real-time strategy games,'' {\em Entertainment Computing}, vol.~52, p.~100843, 2025.

\bibitem{10.1145/3485447.3512083}
X.~Wang, K.~Liu, D.~Wang, L.~Wu, Y.~Fu, and X.~Xie, ``Multi-level recommendation reasoning over knowledge graphs with reinforcement learning,'' in {\em Proceedings of the ACM Web Conference 2022}, WWW '22, (New York, NY, USA), p.~2098–2108, Association for Computing Machinery, 2022.

\end{thebibliography}

\end{document}